\documentclass[runningheads]{llncs}
\usepackage[T1]{fontenc}
\usepackage{hyperref}
\usepackage{booktabs}
\usepackage{comment}

\usepackage{graphicx}
\begin{document}
\title{A FAIR and Free Prompt-based \\ Research Assistant}
%
%
\author{Mahsa Shamsabadi \and Jennifer D'Souza\thanks{Corresponding author: jennifer.dsouza@tib.eu}}
%
\authorrunning{Shamsabadi and D'Souza}
%
\institute{ TIB Leibniz Information Centre for Science and Technology, Hannover, Germany}
\maketitle              
\begin{abstract}

This demo will present the Research Assistant (RA) tool developed to \textit{assist} with six main types of research tasks defined as standardized instruction templates, instantiated with user input, applied finally as prompts to well-known---for their sophisticated natural language processing abilities---AI tools, such as \href{https://chat.openai.com/}{ChatGPT} and \href{https://gemini.google.com/app}{Gemini}. The six research tasks addressed by RA are: creating FAIR research comparisons, ideating research topics, drafting grant applications, writing scientific blogs, aiding preliminary peer reviews, and formulating enhanced literature search queries. RA's reliance on generative AI tools like ChatGPT or Gemini means the same research task assistance can be offered in any scientific discipline. We demonstrate its versatility by sharing RA outputs in Computer Science, Virology, and Climate Science, where the output with the RA tool assistance mirrored that from a domain expert who performed the same research task.


\end{abstract}

\section{Introduction}

\begin{figure}[!htb]
\includegraphics[width=\textwidth]{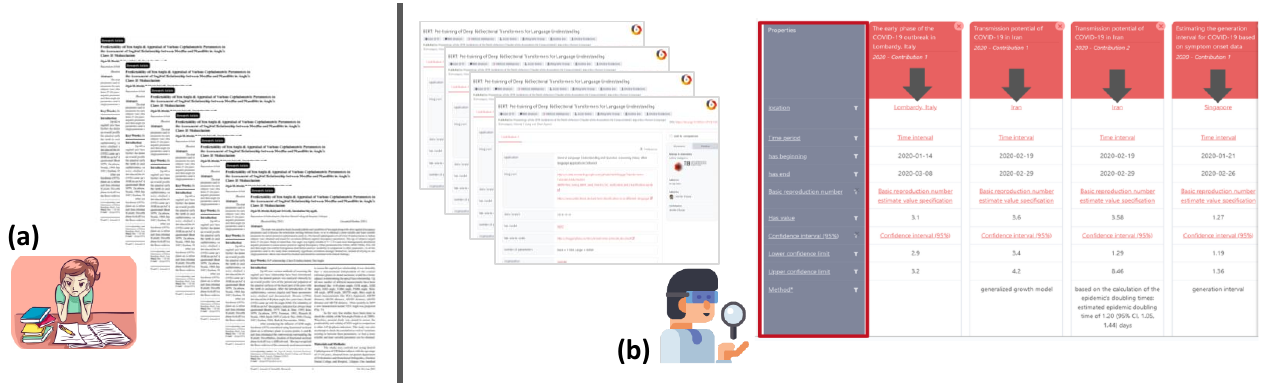}
\caption{In the figure, section (a) presents the traditional discourse-based scholarly communication model where, in the present age of the deluge of publications, the researcher grapples with burdensome cognitive tie-ups including manual note-taking to get a research overview. In contrast, section (b) presents next-generation digital library systems. These systems employ structured representations of machine-actionable scholarly findings, streamlining the comparison process by aggregating similar papers based on shared research dimensions. The grey box indicates common properties, while arrowed columns represent findings from individual scholarly works.} \label{fig:nextgen-orkg}
\vspace{-20pt}
\end{figure}

The shift towards structured, machine-actionable methods of scholarly communication, exemplified by Open Research Knowledge Graph (ORKG) \cite{auer2020improving}, represents a major leap in handling the vast volumes of scholarly communication \cite{bornmann2021growth}. This transition allows for easier comparison of research findings by representing scholarly knowledge as fine-grained, comparable research objects on the semantic web. It ensures adherence to FAIR principles, enhancing accessibility and reusability of scientific knowledge. In contrast, traditional discourse-based methods of scholarly communication are inadequate for researchers to track progress effectively. See Figure \ref{fig:nextgen-orkg} for an illustration.
Emerging AI tools like \href{https://typeset.io/}{SciSpace} and \href{https://elicit.com/}{Elicit} enable automatic extraction of key comparative scientific insights. However, integrating them into FAIR-compliant models like ORKG poses challenges due to their generation of lengthy narrative sentences as comparison values. These verbose outputs are unsuitable as research objects in the semantic web, impeding their assimilation into structured, knowledge graph-based scholarly publishing models like ORKG. Additionally, limited access to these tools is hindered by paywalls, restricting democratization.
Such limitations underscore the need for free access to AI-based research assistants supporting FAIR research comparisons.

In this paper, we introduce a research assistant tool (RA, henceforward) that produces FAIR research knowledge and is accessible free of cost. RA serves to aid scientists in six main research tasks. 1) generating FAIR research comparisons (RC)---as introduced, fine-grained, structured FAIR representations of scholarly knowledge that are readily comparable across their common properties is the next-generation mode of scholarly communication. This is the primary task supported by RA. To this end, RA offers five subtask scenarios for end users: i) comparing entities for a research problem, ii) comparing contexts for a research problem, iii) obtaining salient properties for a research problem, iv) refining definitions of salient properties based on provided context, and v) comparing contexts based on a set of salient properties for a research problem. The next set of four research tasks are secondary and in some cases use the data created in the primary task. They are: 2) brainstorming research ideas (BI). In RA, this is decomposed into two subtasks for end users: i) receiving suggestions for research ideas based on a problem and its properties, or ii) obtaining user stories and criteria based on a context. For instance, based on context discussing climate projections' importance on aquaculture, a user story could be: ``As a stakeholder in marine resource conservation, I want to understand climate-related risks to shellfish aquaculture for effective conservation strategies.'' These stories help identify potential research stakeholders and inform project planning phases where they could be addressed. 3) Writing grant applications (GA)---scientists are also using AI to help write grant applications \cite{owens2023ai}, thus RA assists with preliminary grant writing; 4) generating science blog-posts (BP)---there is an existing impetus to simplify science communication \cite{simpletext2021} from knowledge-dense research papers to informal devices of blog-posts that present the knowledge in simplified form such that it is accessible to lay readers of society, this task is also supported in RA. 5) Writing a preliminary review (PR)---RA addresses the challenges of peer review by helping draft preliminary reviews \cite{shah2022challenges}. And 6) consolidating keyword-based search queries with an exhaustive list of synonyms (KQS)---in composing literature surveys, the initial keyword-based search string is crucial for screening papers from publisher databases \cite{de2023artificial}. For this purpose, RA supports consolidating a boolean search string, i.e. with operators AND, OR, and NOT, by generating an instruction for an LLM to exhaustively add all synonyms for the initial search terms. This reduces user effort in finding synonyms---a task often prone to oversight---, aided, anyway, by existing thesauri. Thus, in total, RA addresses 6 main research tasks, and assists with 11 different research task scenarios. Users of the tool see this as the \href{https://github.com/mahsaSH717/research_assistant/blob/master/screenshots/prompts.png}{initial screen}.


Many see AI conversational agents like \href{https://chat.openai.com/}{ChatGPT} or \href{https://gemini.google.com/app}{Gemini} as tools to assist with work, not replace it, already adopting them as digital secretaries or assistants \cite{owens2023ai}. 
Playing into this familiar setting, engineered into RA underlying each task are a set of customised ChatGPT prompts that assists end users to access relevant AI-generated data that addresses the research task. Thus it standardizes the performance of a diverse spectrum of research tasks, in a single tool, via a modular and completely transparent workflow. RA's FAIRness is evident in its assistance with the first RC research task. Scholarly knowledge once published in the ORKG inherently aligns with FAIR principles. But this is contingent that it is created following the best practices of the semantic web. RA facilitates this by adhering to semantic web's recommended practices for creating structured knowledge, such as prescribing concise resource names and generating definitions for comparison properties. This emphasizes RA's role as a FAIR research assistant, facilitating the creation of FAIR-ready structured knowledge. The last step is publishing the FAIR-ready structured knowledge on the semantic web completing the cycle to meeting the FAIR principles.
The contributions of this work are multi-faceted. 1) RA stands out as the only tool that facilitates the creation of FAIR research comparisons, addressing a critical need in scientific communication. 2) Out of the box, it introduces 11 standardized ChatGPT prompts tailored to specific research tasks, showcasing a broadened scope of possibilities that NLP researchers can consider in developing new applications using conversational AI in research. In contrast to existing AI tools like \href{https://elicit.com/}{Elicit} or \href{https://typeset.io/}{SciSpace}, it offers a novel perspective on AI's application in research practices. 3) RA's workflow is transparent and modularly designed. With the public release of its source code \url{https://github.com/mahsaSH717/research_assistant}, users can effortlessly expand RA by engineering new research tasks. 

\section{The Research Assistant Tool---A Walkthrough of its Research Tasks' Assistance}

\begin{figure}[!htb]
\includegraphics[width=\textwidth]{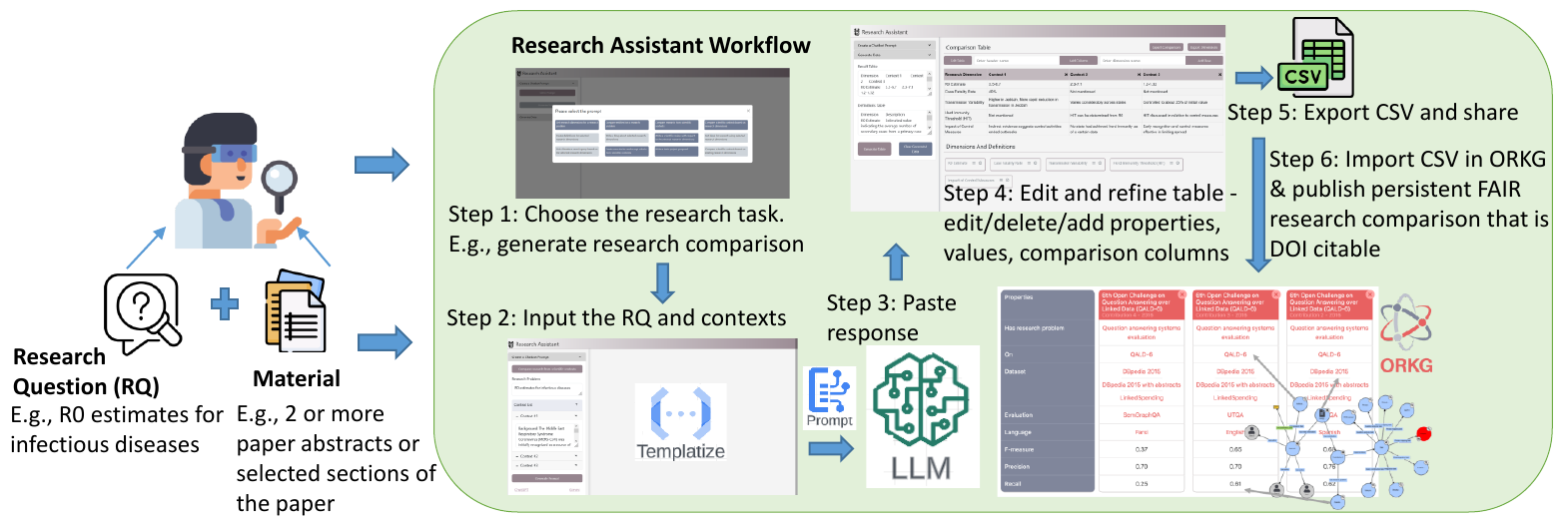}
\caption{The Research Assistant (RA) simplifies the workflow for generating research comparisons based on user input of a research question and selected research contexts. RA provides a detailed ChatGPT prompt to perform the task, resulting in a table that offers the desired research overview. Once generated, a research comparison can also be repurposed as a dataset for other researchers interested in the same research question. RA supports post-editing, thus allowing for pasting the LLM response back in the interface where improper facets of the automatically generated comparison can be curated to make it gold-standard ready which, in a final step, can be exported from RA as a \href{https://github.com/mahsaSH717/research_assistant/blob/master/examples/ORKG\%20Comparison\%20Files/research-assistant-final-comparison.csv}{csv file} shareable via their preferred data repository or \href{https://orkg.org/comparison/R675568/}{published persistently} FAIR, via its \href{https://github.com/mahsaSH717/research_assistant/blob/master/examples/ORKG\%20Comparison\%20Files/import-file.csv}{direct csv import} feature, on the Open Research Knowledge Graph (ORKG).} \label{fig:result-rc}
\vspace{-20pt}
\end{figure}



We use a scenario to demonstrate the use of RA. \textit{Sarah} is a fresh undergraduate student in Computer Science and is looking for the research frontier around GPT series large language models. Specifically, she needs to identify what their salient properties are and compare them based on their salient properties. She encounters RA and starts up the tool. There in she is faced with the option of 11 tasks (depicted  \href{https://github.com/mahsaSH717/research_assistant/blob/master/screenshots/prompts.png}{here}). She begins by selecting the ``Get research dimensions for research problem'' task in RA, entering ``GPT family of large language models'' as the research problem. Given this input RA generates a ChatGPT prompt which generates the following list of properties with their descriptions: architecture, training data, model size, fine-tuning, evalaution metrics and \href{https://chat.openai.com/share/22c5a0e9-3cd5-41e3-817a-7864b52eb1d5}{more}. This gives a starting systematic view on the problem. She can paste this response back into RA. Now she wishes to have a comparative view of the models themselves and she has the context from the respective model's papers. She can select the research task ``Compare scientific contexts based on research dimensions.'' She now has the option to paste the contexts into RA. Reusing the earlier generated properties and the contexts, RA generates a new ChatGPT prompt. The response she gets is a \href{https://chat.openai.com/share/fc1af53d-be88-4a73-93ab-71796a79db17}{systematic comparison} of GPT 1, 2, and 3. Consider that Sarah now has a quick overview of the models in a matter of a few minutes. The traditional scenario would be that Sarah would have to spend a day(s) reading each of the papers just to get this overview. Now equipped with a quick overview, Sarah can focus on specific details of interest in the papers. Then she decides that she wants to persistently publish this overview to other researchers interested in the same question without the need for them to repeat the same task. RA supports post-editing functions, allowing her to customize and edit the tables before exporting the data as a csv file. She can then publish it on the ORKG which is a next-generation digital library for publishing FAIR research comparisons. She now has a persistent citeable DOI for her comparison which she can share in the community. Later, Sarah decides she wants to write a blog-post on her website on the GPT family of models discussing some of their properties. Within RA, she selects the properties she wants to discuss, and RA generates a ChatGPT prompt for her. Based on the \href{https://chat.openai.com/share/d750d275-c16d-4b22-a88d-49bee5c66ae1}{response}, she can choose to edit it in her favorite text editor and publish it as descriptive piece in non-scientific language on the problem. Lastly, Sarah seeks ideas for further research that she can explore. In RA, she can select the task ``Get ideas for research using selected research dimensions'', then all she needs to do is select the set of properties of the GPT models that piqued her interest. RA will then generate a ChatGPT prompt for her. The LLM can then be put to the task of suggesting \href{https://chat.openai.com/share/88674b65-efb3-4662-8d41-a3afff14230f}{research ideas}. \autoref{fig:result-rc} shows the workflow of RA to generate FAIR research comparisons.

One of the immediate benefits of RA is that it offers domain-independent research assistance. \autoref{tab1} shows examples of three different scientific domains, i.e. Computer Science, Virology, and Climate Science, where we tested RA respectively. Moreover, we chose specific research problems within each of the domains. For each research problem, while each of the 11 scenarios were visited with the tool, on a surface look at the research dimensions in the primary application of RA, we find the LLM output to be overlapping and complementary to the human curated data in the ORKG seemingly a very useful assistant. Detailed walkthroughs of the RA usage scenarios for three different scientific research problems can be found in the \href{https://github.com/mahsaSH717/research_assistant/tree/master/examples}{examples folder} in our codebase.

\begin{table}[!htb]
\vspace{-5mm}
\caption{Three research-problem-based use-cases of the Research Assistant (RA) tool.}\label{tab1}
\begin{tabular}{|p{3cm}|p{4.5cm}|p{4.5cm}|} \toprule
Research problem & Human-curated dimensions & LLM-generated dimensions  \\ \midrule
A catalog of transformer models & model name, model family, date created, pre-training architecture, pre-training task, training corpus, optimizer, \href{https://orkg.org/comparison/R609337/}{more here} & parameters, architecture, pre-training data, model size, vocabulary size, layer configuration, optimizer, \href{https://github.com/mahsaSH717/research_assistant/tree/master/examples/LLMS}{more here} \\ \hline
R0 estimates for infectious diseases & disease name, location, R0 value, time period, \href{https://orkg.org/comparison/R44930/}{more here} & R0 estimate, incubation period, zoonotic origin, \href{https://github.com/mahsaSH717/research_assistant/tree/master/examples/R0Estimates}{more here} \\ \hline
Prediction of impact of climate change on species & species, climate exposure, climate scenario, world region, impact, \href{https://orkg.org/comparison/R583761/}{more here} & impact on species, geographical range, adaptation potential, biodiversity impact, \href{https://github.com/mahsaSH717/research_assistant/tree/master/examples/ImpactOfClimateChange}{more here} \\ \bottomrule
\hline
\end{tabular}
\vspace{-10mm}
\end{table}

\subsection{Implementation Technical Details}

The RA tool is a \href{https://react.dev/}{React}-based front-end application created with \href{https://create-react-app.dev/}{Create React App}, which sets up an initial React application. The application has four main custom components and each might contain smaller child components (see \url{https://react.dev/reference/react/components}). The first component displays a \href{https://github.com/mahsaSH717/research_assistant/blob/master/screenshots/prompts.png}{list of prompts}, while the second handles input fields for selected prompts, activating the 'Generate Prompt' button once inputs are provided. Consider this \href{https://github.com/mahsaSH717/research_assistant/blob/master/screenshots/input.png}{screenshot}. The third component is responsible for generating comparison tables from conversational agent data in the 'Generate Data' area. The fourth component displays these tables and definitions on the right side. See \href{https://github.com/mahsaSH717/research_assistant/blob/master/screenshots/results.png}{screenshot}. The project incorporates several third-party libraries, including \href{https://react-bootstrap.netlify.app/}{React Bootstrap} for UI component, \href{https://github.com/nkbt/react-copy-to-clipboard}{react-copy-to-clipboard} for copying prompts, \href{https://github.com/react-csv/react-csv}{react-csv} for exporting tables and definitions as CSV files, and \href{https://react-icons.github.io/react-icons/}{React Icons} for graphical elements. Moreover, we used \href{https://github.com/css-modules/css-modules}{CSS Modules} which allows component-scoped styling to style the components in this project.

\section{Conclusions}

In conclusion, RA offers a user-centric approach to streamlining the creation of FAIR research comparisons, as its primary task, while ensuring data control for users. With data stored directly on users' devices and accessible through an intuitive interface compatible with Excel, RA simplifies data management, updates, and sharing. The platform's csv export feature facilitates seamless integration of comparisons into next-generation digital libraries like the ORKG, ensuring the insights are reshareable and persistently citeable. In a broader sense, by addressing a diverse spectrum of research tasks, RA aims to leverage unhindered assistance from AI. Customized research-task-specific ChatGPT prompts enable users to access relevant AI-generated data efficiently. By capitalizing on the familiarity with conversational agents like ChatGPT and Gemini, RA positions itself as an indispensable research assistant across the board in Science. We propose RA as a demonstration application  for an on-site usability study. AI-based scientific tools, despite their potential, sometimes are underused due to usability issues, steep learning curves, misalignment research methods, and distrust in their reliability \cite{van2019usage}. As such, by showcasing the RA tool to the NLDB attendees, we plan a survey for usability factors and application scenarios.



\section*{Acknowledgements}

This work was supported by the German BMBF project SCINEXT (ID 01lS22070).

%
%
%
\bibliographystyle{splncs04}
\bibliography{mybibliography}

\end{document}